\documentclass[runningheads]{llncs}

\usepackage{comment}

\usepackage{float}
\usepackage{color}
\usepackage{epsfig}
\usepackage{graphicx}
\usepackage{amsmath}
\usepackage{amssymb}
\usepackage{url}
\usepackage[breaklinks=true,bookmarks=false]{hyperref}

\usepackage{subcaption}
\usepackage{xcolor}
\usepackage{multirow}
\hypersetup{
	hidelinks,
	colorlinks=true,
}
\usepackage{microtype}
\usepackage{enumitem}

\newcommand{\loss}{\mathcal{L}}
\newcommand{\totalloss}{\loss_{total}}
\newcommand{\berhuloss}{\loss_{B}}
\newcommand{\gradientloss}{\loss_{g}}
\newcommand{\normalloss}{\loss_{n}}
\newcommand{\relativeloss}{\loss_{r}}
\newcommand{\krelativeloss}{\loss_{r,k}}
\newcommand{\pixel}{p}
\newcommand{\accuracy}{\delta}
\newcommand{\model}{\mathcal{M}}
\newcommand{\baselinemodel}{\model_{b}}
\newcommand{\edgeawaremodel}{\model_{e}}
\newcommand{\sabmodel}{\model_{e,SAB}}
\newcommand{\relativelossmodel}{\model_{e,\relativeloss}}
\newcommand{\fullmodel}{\model_{full}}

\def\etal{\emph{et al.}}

\newcommand{\nothing}[1]{}

\definecolor{TabGreenColor}{rgb}{0,0.5,0}
\definecolor{TabBlueColor}{rgb}{0,0.5,0.8}

\graphicspath{{./image/}}
\begin{document}



\title{Improving Monocular Depth Estimation by Leveraging Structural Awareness and Complementary Datasets} 

\titlerunning{Improving Monocular Depth Estimation}

\newcommand*\samethanks[1][\value{footnote}]{\footnotemark[#1]}

\author{
Tian Chen\thanks{Joint first authors} \and
Shijie An\samethanks \and
Yuan Zhang \and
Chongyang Ma\ \and
\\
Huayan Wang \and
Xiaoyan Guo \and
Wen Zheng
}
%
\authorrunning{Chen \emph{et al.}}
%

\institute{
Y-tech, Kuaishou Technology}

\maketitle


\begin{abstract}
Monocular depth estimation plays a crucial role in 3D recognition and understanding. One key limitation of existing approaches lies in their lack of structural information exploitation, which leads to inaccurate spatial layout, discontinuous surface, and ambiguous boundaries. In this paper, we tackle this problem in three aspects. First, to exploit the spatial relationship of visual features, we propose a structure-aware neural network with spatial attention blocks. These blocks guide the network attention to global structures or local details across different feature layers. Second, we introduce a global focal relative loss for uniform point pairs to enhance spatial constraint in the prediction, and explicitly increase the penalty on errors in depth-wise discontinuous regions, which helps preserve the sharpness of estimation results. Finally, based on analysis of failure cases for prior methods, we collect a new Hard Case (HC) Depth dataset of challenging scenes, such as special lighting conditions, dynamic objects, and tilted camera angles. The new dataset is leveraged by an informed learning curriculum that mixes training examples incrementally to handle diverse data distributions. Experimental results show that our method outperforms state-of-the-art approaches by a large margin in terms of both prediction accuracy on NYUDv2 dataset and generalization performance on unseen datasets.
\end{abstract}

\section{Introduction}
\label{sec:intro}
Recovering 3D information from 2D images is one of the most fundamental tasks in computer vision with many practical usage scenarios, such as object localization, scene understanding, and augmented reality. Effective depth estimation for a single image is usually desirable or even required when no additional signal (e.g., camera motion and depth sensor) is available.
However, monocular depth estimation (MDE) is well known to be ill-posed due to the many-to-one mapping from 3D to 2D.
To address this inherent ambiguity, one possibility is to leverage auxiliary prior information, such as texture cues, object sizes and locations, as well as occlusive and perspective clues~\cite{saxena2006learning,karsch2014depth,ladicky2014pulling}.

More recently, advances in deep convolutional neural network (CNN) have demonstrated superior performance for MDE by capturing these priors  implicitly and learning from large-scale dataset~\cite{eigen2014depth,eigen2015predicting,laina2016deeper,xu2017multi,qi2018geonet,hu2019revisiting}.
CNNs often formulate MDE as classification or regression from pixels values without explicitly accounting for global structure. That leads to loss of precision in many cases. To this end, we focus on improving structure awareness in
monocular depth estimation.

\begin{figure*}[t]
	\centering
	\begin{minipage}{.48\linewidth}\centering
		\includegraphics[width=0.99\linewidth]{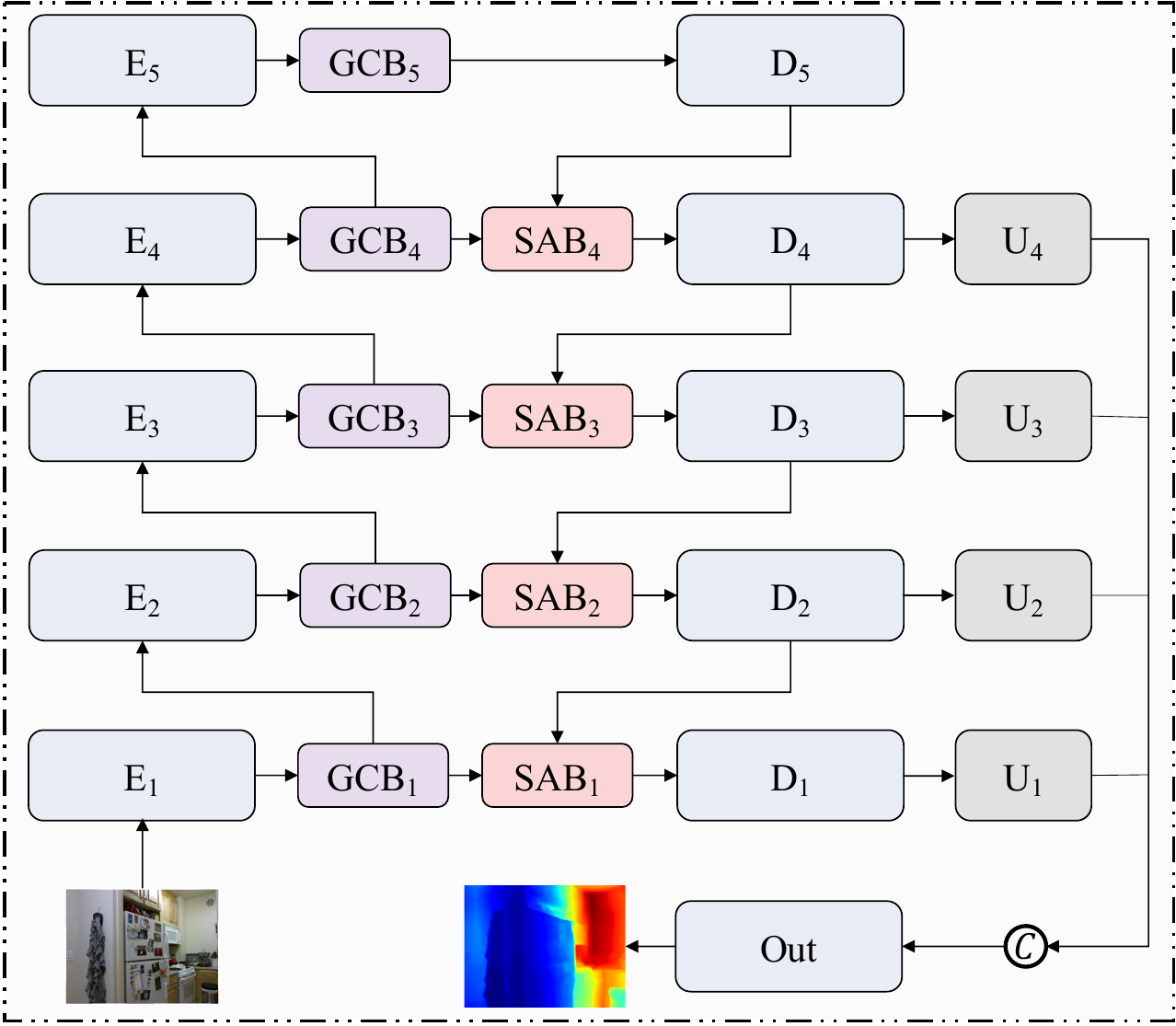}
		\subcaption{Our network architecture.
		$\text{E}_i$/$\text{D}_i$ denotes the $i$-th encoder/decoder\nothing{ while the index $i$ denotes the scale of a feature map}.
		$\text{SAB}_i$ module couples a high-level feature from $D_{i+1}$ with a low-layer feature from $\text{GCB}_i$\nothing{ via a skip connection}.
		\textcircled{\centering \textit{c}} is the concatenate operation for the output of upsampling blocks $\{ \text{U}_i\}$. }
	\end{minipage}
	\quad
	\begin{minipage}{.48\linewidth}\centering
		\begin{minipage}{.99\linewidth}\centering
			\includegraphics[width=0.99\linewidth]{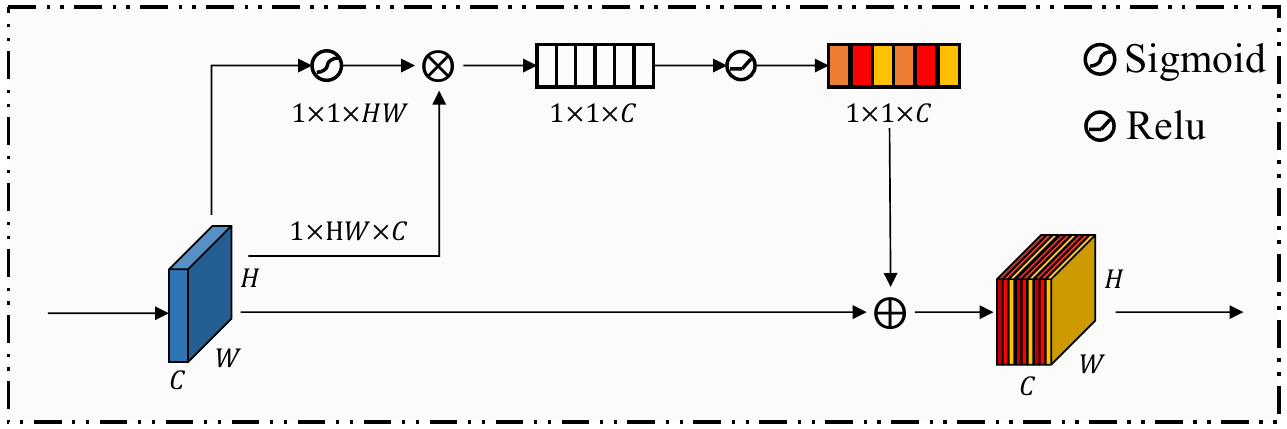}
			\subcaption{Global context block (GCB)~\cite{cao2019gcnet}.
			The feature maps are shown as the shape of their tensors, \textit{e.g., $H \times W \times C$}\nothing{ for $C$ channels}.
			$\otimes$ denotes matrix multiplication and $\oplus$ denotes element-wise sum. }
		\end{minipage}
		\begin{minipage}{.99\linewidth}\centering
			\includegraphics[width=0.99\linewidth]{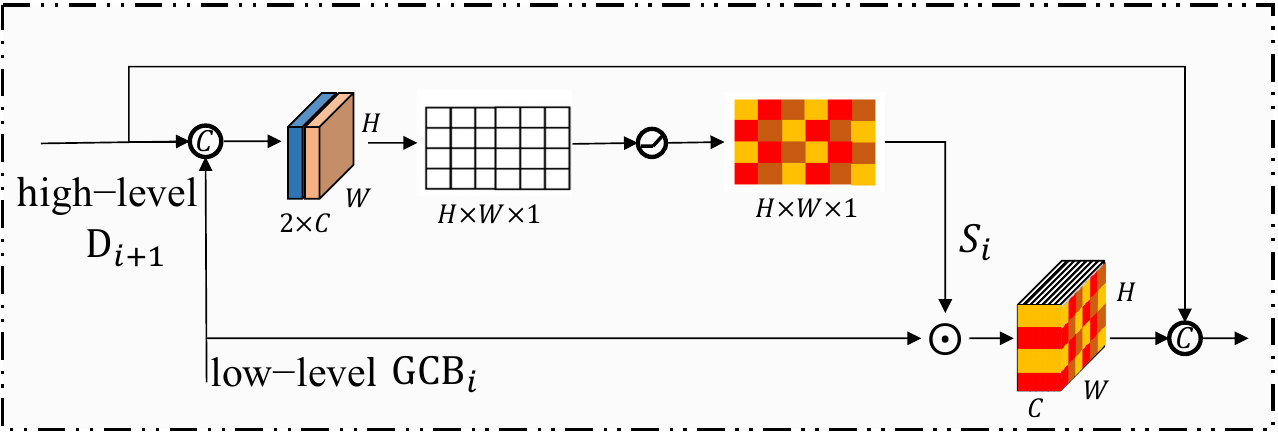}
			\subcaption{Our spatial attention block (SAB).
			$S_i$ represents the attention map of $\text{SAB}_i$ and $\odot$ denotes element-wise dot product.}
		\end{minipage}
	\end{minipage}
	\caption{Illustration of our network architecture and the proposed SAB.}
	\label{fig_method_structure}
\end{figure*}

Specifically, we propose a new network module, named \emph{spatial attention block}, which extracts features via blending cross-channel information. We sequentially adopt this module at different scales in the decoding stage (as shown in Fig.~\ref{fig_method_structure}a) to generate spatial attention maps which correspond to different levels of detail.
We also add a novel loss term, named \emph{global focal relative loss} (GFRL), to ensure sampled point pairs are ordered correctly in depth.
Although existing methods attempt to improve the visual consistency between predicted depth and the RGB input, they typically lack the ability to boost performance in border areas, which leads to a large portion of quantitative error and inaccurate qualitative details\nothing{, which calls for a more targeted solution}.
We demonstrate that simply assigning larger weights to edge areas in the loss function can address this issue effectively.

\begin{figure}[t]
 	\centering
 	\includegraphics[width=.9\linewidth]{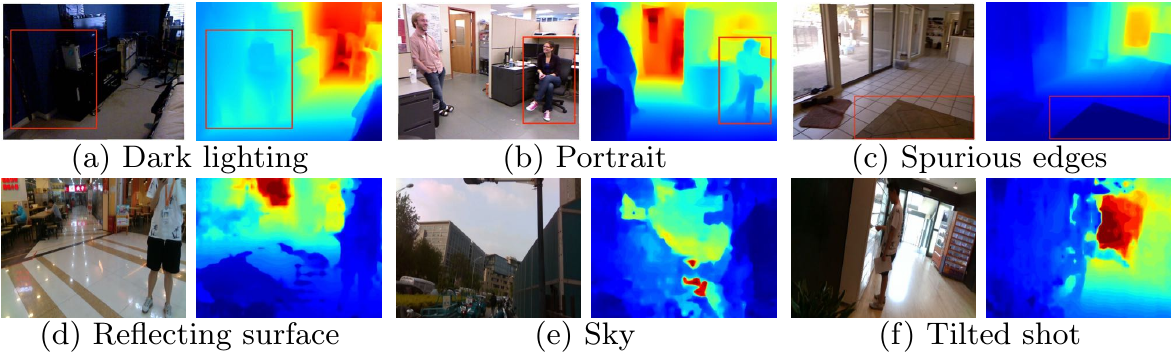}
	\caption{Six typical hard cases for existing monocular depth estimation methods. (a), (c), and (e) show results of Alhashim \etal~\cite{Alhashim2018}, while (b), (d), and (f) are based on Fu \etal~\cite{fu2018deep}. {\color{red} Red} boxes highlight inaccurate regions in the results.}
	\label{fig_abstract_HC}
 \end{figure}

Furthermore, MDE through CNNs usually cannot generalize well to unseen scenarios~\cite{van2019neural}. We find six types of common failure cases as shown in Fig.~\ref{fig_abstract_HC} and note that the primary reason for these failures is the lack of training data, even if we train our network on five commonly used MDE datasets combined\nothing{ which provide high-resolution depth maps}.
To this end, we collect a new dataset, named \emph{HC Depth Dataset}, to better cover these difficult cases. 
We also show that an incremental dataset mixing strategy inspired by curriculum learning can improve the convergence of training when we use data following diverse distributions.

To sum up, our main contributions include:
\begin{itemize}
\item A novel spatial attention block in the network architecture.
\item A new loss term (GFRL) and an edge-aware consistency scheme. 
\item A new MDE dataset featuring hard cases that are missing or insufficient in existing datasets, and a data mixing strategy for network training. 
\end{itemize}

\section{Related Work}

\paragraph{Monocular depth estimation.}
Depth estimation from 2D images is an essential step for 3D reconstruction, recognition, and understanding.
Early methods for depth estimation are dominated by geometry-based algorithms which build feature correspondences between input images and reconstruct 3D points via triangulation~\cite{hirschmuller2008stereo,khamis2018stereonet}.
Recently CNN-based approaches for pixel-wise depth prediction~\cite{kong2019pixel,eigen2015predicting,xu2017multi} present promising results from a single RGB input, based on supervision with ground-truth training data collected from depth sensors such as LiDAR and Microsoft Kinect camera.
By leveraging multi-level contextual and structural information from neural network, depth estimation has achieved very encouraging results~\cite{garg2016unsupervised,kuznietsov2017semi,laina2016deeper,yin2019enforcing}.
The major limitation of this kind of methods is that repeated pooling operations in deep feature extractors quickly decrease the spatial resolution of feature maps. To incorporate long-range cues which are lost in downsampling operations, a variety of approaches adopt skip connections to fuse low-level depth maps in encoder layers with high-level ones in decoder layers~\cite{kuznietsov2017semi,fu2018deep,xu2017multi}.

Instead of solely estimating depth, several recent multi-task techniques~\cite{eigen2015predicting,qi2018geonet,hu2019revisiting} predict depth map together with other information from a single image.
These methods have shown that the depth, normal, and class label information can be jointly and consistently transformed with each other in local areas.
However, most of these approaches only consider local geometric properties, while ignoring global constraints on the spatial layout and the relationship between individual objects.
The most relevant prior methods to ours are weakly-supervised approaches which consider global relative constraint and use pair-wise ranking information to estimate and compare depth values~\cite{chen2016single,xian2018monocular,chen2019learning}.

\paragraph{Attention mechanism.}
Attention mechanisms has been successfully applied to various high-level tasks, such as generative modeling, visual recognition, and object detection~\cite{zhang2018self,wang2017residual,hu2018relation}.
In addition, attention maps are very useful in pixel-wise tasks.
NLNet~\cite{wang2018non} adopts self-attention mechanism to model the pixel-level pairwise relationship.
CCNet~\cite{huang2018ccnet} accelerates NLNet by stacking two criss-cross blocks, which extract contextual information of the surrounding pixels.
Yin~\etal~\cite{xu2018structured} leverage multi-scale structured attention model which automatically regulates information transferred between corresponding features.

\paragraph{Cross-dataset knowledge transfer.}
A model trained on one specific dataset generally does not perform well on others due to dataset bias~\cite{torralba2011unbiased}.
For MDE, solving different cases, \emph{e.g.}~indoor, outdoor, and wild scenes, usually requires explicitly training on diverse datasets~\cite{li2019learning,chen2019learning,gordon2019depth,Lasinger2019}.
When training on mixed datasets, curriculum learning~\cite{bengio2009curriculum} is needed to avoid the local minimum problem by training the model on \emph{easier} datasets first.
Moreover, when the datasets are imbalanced, resampling~\cite{he2009learning,he2013imbalanced,oquab2014learning} is often performed to reshape the data distribution.

\section{Our Method}
\label{chap_method}
\label{sec:method}

\nothing{
This section describes our MDE method.
The method demonstrates good performance through our four aspects:
a novel attention-based encoder-decoder architecture as shown in Fig.~\ref{fig_method_structure}(a);
a new proposed global constraint loss function for MDE;
a depth edge focusing strategy to improve prediction of details;
a new dataset---HC Depth dataset which is utilized to improve the performance of the network by incremental mixing strategy.
}

\subsection{Network Architecture}
We illustrate our network architecture in Fig.~\ref{fig_method_structure}a.
Based on a U-shaped network with an encoder-decoder architecture~\cite{mayer2016large}, we add skip connections~\cite{wang2015towards} from encoders to decoders for multi-level feature maps. We observe that encoder mainly extracts semantic features, while decoder pays more attention to spatial information.
A light-weight attention based global context block (GCB)~\cite{cao2019gcnet} (Fig.~\ref{fig_method_structure}b) is sequentially applied to each residual block in the encoding stage to recalibrate channel-wise features.
The recalibrated vectors with global context are then combined with high-level features as input for our spatial attention blocks (SAB) through skip connections.
Then, the recalibrated features and the high-level features are fused to make the network focus on either the global spatial information or the detail structure at different stages.
From a spatial point of view, the channel attention modules are applied to emphasize semantic information \emph{globally}, while the spatial attention modules focus on where to emphasize or suppress \emph{locally}.
With skip connections, these two types of attention blocks build a 3D attention map to guide feature selection.
The output features of the four upsampling blocks are upsampled by a factor of $2$, $4$, $8$, and $16$, respectively, and then fed to the refinement module to obtain the final depth map which has the same size as the original image.
The main purpose of these multi-scale output layers is to fuse information at multiple scales together.
The low resolution output retains information with finer global layout, while the high resolution output is used to restore details lost after the downsampling operations.


\paragraph{Spatial attention block.}
Different from channel-attention mechanism which selects semantic feature in network derivation, our SAB is designed to optimize geometric spatial layout in pixel-wise regression tasks.
In SAB, we perform a squeeze operation on concatenate features via $1\times1$ convolution to aggregate spatial context across their channel dimensions. Then we activate local attention to get a 2D attention map which encodes pixel-wise depth information over all spatial locations.
The low-level features are multiplied by this 2D attention map for subsequent fusion to deliver the spatial context from a higher-level layer.
Therefore, our SAB generates attention maps with richer spatial information to recalibrate the semantic features from GCB.
The SAB shown in Fig.~\ref{fig_method_structure}(c) can be formulated as
\begin{equation}
\text{D}_{i}=f\Big(\sigma\big(W_{1}*f(\text{D}_{i+1}, \text{GCB}_{i})\big)\odot \text{GCB}_{i}, \text{D}_{i+1}\Big),
\end{equation}
where $f$ is a fusion function (e.g. element-wise sum, element-wise dot product, or concatenation), while $*$ denotes $1\times1$ or $3\times3$ convolution and $\odot$ denotes element-wise dot product.
Since a depth maps has a wide range of positive values, we use $\mathrm{ReLU}$ as the activation function $\sigma(x)$.

\begin{figure}
	\centering
	\includegraphics[width=.9\linewidth]{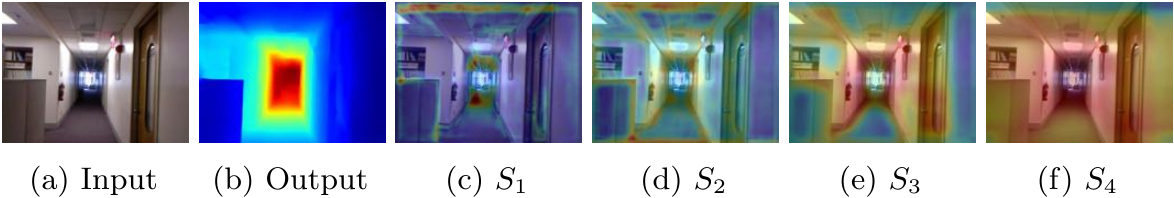}
	\caption{Spatial attention maps at different scales visualized as heat maps overlaid on the input image. $S_i$ is the $i$th attention map from the top layer in the decoder (see Fig.~\ref{fig_method_structure}c). Red color indicates larger value. Our network attends to different levels of structural detail at different scales.}
	\label{fig_method_DFG}
\end{figure}

As shown in Fig.~\ref{fig_method_DFG}, attention feature maps obtained using our SAB help the network focus on the specific information of relative depth across different levels.
Specifically, the attention map $S_4$ contains low-level features which depict the \nothing{partial }semantic hierarchy and capture the overall near-and-far structure in 3D space.
The closer a spatial attention feature map is to the top layer $S_1$, the more local details are focused for the prediction output.

\subsection{Network Training}
\label{chap_loss}
\label{sec:network_training}

The loss function to train our network contains four terms, i.e., Berhu loss $\berhuloss$, scale-invariant gradient loss $\gradientloss$, normal loss $\normalloss$, and global focal relative loss $\relativeloss$.
We describe each loss term in detail as follows.

\paragraph{BerHu loss.}
The BerHu loss refers to the reversed Huber penalty~\cite{zwald2012berhu} of depth residuals, which provides a reasonable trade-off between L1 norm and L2 norm in regression tasks when the errors present a heavy-tailed distribution~\cite{roy2016monocular,laina2016deeper}.
Therefore, the BerHu loss $\berhuloss$  is commonly used as the basic error metric for MDE and is defined as:
\begin{equation}
	\berhuloss =\sum_{i, j}|d_{i, j}-\hat{d}_{i, j}|_{b}, ~|x|_{b}  = \begin{cases}
		\left| x\right|  & \left| x\right| \leq c,  \\
		\frac{x^2 +c^2}{2c}  &  \left| x\right| > c\end{cases},
\end{equation}
where $d_{i, j}$ and $\hat{d}_{i, j}$ are the ground-truth and predicted depth values at the pixel location $(i, j)$, respectively.
We set $c$ to be $0.2\max\limits_{p}(|\hat{d}_{p}-d_{p}|)$, where $\{ p \}$ indicate all the pixels\nothing{ on each image} in one batch.

\paragraph{Scale-invariant gradient loss.}
We use scale-invariant gradient loss~\cite{ummenhofer2017demon} to emphasize depth discontinuities at object boundaries\nothing{, stimulate sharp edges in the depth map,} and to improve smoothness in homogeneous regions.
To cover gradients at different scales in our model, we use $5$ different spacings $\{ s = 1,2,4,8,16 \}$ for this loss term $\gradientloss$:
\begin{equation}
\begin{aligned}
\gradientloss&=\sum_{s} \sum_{i, j}|\mathbf{g}_{s}(i, j)-\mathbf{\hat{g}}_{s}(i, j)|^{2}, \\
\mathbf{g}_{s}(i, j)&=\left(\frac{d_{i+s,j}-d_{i,j}}{|d_{i+s,j}+d_{i, j}|}, \frac{d_{i, j+s}-d_{i, j}}{|d_{i, j+s}+d_{i, j}|} \right)^{\top},
\end{aligned}
\end{equation}

\paragraph{Normal loss.}
To deal with small-scale structures and to further improve high-frequency details in the predicted depth, we also use a normal loss term $\normalloss$:
\begin{equation}
\normalloss=\sum_{i, j}\left(1-\frac{\left\langle \mathbf{n}_{i, j}, \mathbf{\hat{n}}_{i, j}\right\rangle}{\sqrt{\left\langle \mathbf{n}_{i, j},
\mathbf{n}_{i, j}\right\rangle} \cdot \sqrt{\left\langle \mathbf{\hat{n}}_{i, j}, \mathbf{\hat{n}}_{i, j}\right\rangle}}\right),
\end{equation}
in which $\langle\cdot, \cdot\rangle$ denotes the inner product of two vectors.
The surface normal is denoted as $\mathbf{n}_{i,j}=\left[-\nabla_{x},-\nabla_{y}, 1\right]^{\top}$, where $\nabla_{x}$ and $\nabla_{y}$ are gradient vectors along the $x$ and $y$-axis in the depth map, respectively.

\paragraph{Global focal relative loss.}
Relative loss (RL)~\cite{chen2016single,xian2018monocular,lee2019monocular,chen2019learning} is used to make depth-wise ordinal relations between sample pairs predicted by the network consistent with the ground-truth\nothing{,
but cannot guarantee uniform selection of point pairs and tends to extend them infinitely}.
Inspired by the focal loss defined in Lin \etal~\cite{lin2017focal}, we propose an improved version of relative loss, named \emph{global focal relative loss} (GFRL), to \nothing{adaptively down-weight these pairs and thus }put more weight on sample pairs of incorrect ordinal relationships in the prediction.
To ensure uniform selection of point pairs, we subdivide the image into $16\times16$ blocks of the same size and randomly sample one point from each block.
Each point is compared with all the other points from the same image when training the network.
Our loss term of $n$ pairs is formally defined as $\relativeloss=\sum^n_k\krelativeloss$.
The $k$-th pair loss term $\krelativeloss$ is
\begin{equation}
\begin{aligned}
\krelativeloss &=
\begin{cases} w_k^{\gamma} \log \Big(1+\exp \big(-r_k \left(d_{1,k}-d_{2,k}\right)\big)\Big), & r_{k} \neq 0
\\ \left(d_{1,k}-d_{2,k}\right)^{2}, & r_k=0
\end{cases}
\\
w_k &=1-1 / \big(1+\exp \big(-r_k \left(d_{1,k}-d_{2,k}\right)\big)\big),
\label{eqn:relative_loss}
\end{aligned}
\end{equation}
where $r_k$ is the ground-truth ordinal relationship and is set to $-1$, $0$, or $1$, if the first point has a smaller, equal, or a larger depth value compared to the second point.
The equality holds if and only if the depth difference ratio is smaller than a threshold of $0.02$.

In Eqn.~\ref{eqn:relative_loss}, our key idea is to introduce a modulating factor $w_{k}^{\gamma}$ to scale the relative loss. 
Intuitively, when a pair of pixels have incorrect ordinal relationship in the prediction, the loss is unaffected since $w_{k}$ is close to $1$.
If the depth ordinal relationship is correct and the depth difference is large enough, the weight $w_{k}$ on this pair will go to $0$.
The parameter $\gamma$ smoothly adjusts the magnitude of weight reduction on easy point pairs.
When $\gamma= 0$, our GFRL is equivalent to RL~\cite{chen2016single}.
As $\gamma$ increases, the impact of the modulating factor becomes larger and we set $\gamma= 2$ in our experiments.
It turns out our GFRL outperforms RL under various evaluation metrics (see Sec.~\ref{sec:experimental_results}).

\paragraph{Total loss.}
We train our network in a sequential fashion for better convergence by using different combinations of loss terms in different stages~\cite{hu2019revisiting}.
Our total loss function $\totalloss$ is defined as:
\begin{equation}
	\totalloss=
	\lambda_1\berhuloss^{\text{\uppercase\expandafter{\romannumeral1}-\uppercase\expandafter{\romannumeral3}}} +
	\lambda_2\gradientloss^{\text{\uppercase\expandafter{\romannumeral2}-\uppercase\expandafter{\romannumeral3}}}+
	\lambda_3\normalloss^{\text{\uppercase\expandafter{\romannumeral3}}}+
	\lambda_4\relativeloss^{\text{\uppercase\expandafter{\romannumeral3}}},
\end{equation}
where $\{ \lambda_i \}$ are the weights of different loss terms, and the superscripts $\text{\uppercase\expandafter{\romannumeral1}-\uppercase\expandafter{\romannumeral3}}$ denote the stages of using the corresponding terms.
BerHu loss is the basic term being used to start the training.
After convergence, we first add gradient loss for smooth surface and sharp edges.
To further improve the details of predicted depth and refine the spatial structure, we add normal loss and global focal relative loss in the final stage.

\paragraph{Edge-aware consistency.}
\nothing{It is challenging to predict the depth of object boundaries accurately in an MDE task.
These boundaries can be categorized into \emph{depth boundaries} (e.g., the desk boundary in the first row of Fig.~\ref{fig_method_boundary}) where the the depth values should be discontinuous, and \emph{color boundaries} (e.g., the fresco on the wall in the second row of Fig.~\ref{fig_method_boundary}) where the depth map should be continuous.
Existing MDE methods tend to confuse these boundaries as shown in the case of spurious edges in Fig.~\ref{fig_abstract_HC}.
Besides, we observe that the average prediction error of boundary areas is about $4$ times larger than that of other areas.
To preserve depth boundaries and to decrease discontinuities in regions of color boundaries, we introduce edge-aware consistency to improve the depth estimation accuracy for our method.
Specifically, we first use Canny edge detector~\cite{bao2005canny} to extract edges for the ground-truth depth map, and then dilate these edges with a kernel of $5$ to get a boundary mask (see the right column in Fig.~\ref{fig_method_boundary}).
We multiply the loss $\loss_{ij}$ at the pixel $\pixel_{ij}$ by a weight of $5$ if $\pixel_{ij}$ is an edge pixel in the boundary mask.}

Depth discontinuities typically arise at the boundaries of occluding objects in a scene.
Most existing MDE methods cannot recover these edges accurately and tend to generate distorted and blurry object boundaries~\cite{hu2019revisiting}.
Besides, we observe that in these areas, the average prediction error is about $4$ times larger than that of other areas.

Based on these observations, we introduce an edge-aware consistency scheme to preserve sharp discontinuities in depth prediction results.
Specifically, we first use Canny edge detector~\cite{bao2005canny} to extract edges for the ground-truth depth map, and then dilate these edges with a kernel of $5$ to get a boundary mask.\nothing{(see the right column in Fig.~\ref{fig_result_boundary})}
We multiply the loss $\loss_{ij}$ at the pixel $\pixel_{ij}$ by a weight of $5$ if $\pixel_{ij}$ is an edge pixel in the boundary mask.
Our edge-aware consistency scheme can be considered as a hard example mining method~\cite{shrivastava2016training} to explicitly increase the penalty on prediction errors in boundary regions.
\nothing{
We set the depth edge areas as a mask with a weight of $5$, while the weight of other areas is set to $1$:
\begin{equation}
\mathcal{L}^e_{ij}=w_{ij}\mathcal{L}_{ij}, \quad
w_{ij}=\begin{cases}
5,\quad(i,j)\in\Gamma\\
1,\quad(i,j)\notin\Gamma
\end{cases},
\end{equation}
where $\Gamma$ is the set of edge pixels and $\mathcal{L}_{i j}^{e}$ is the final loss at the pixel $p(i,j)$ weighted by $w_{ij}$.
}

\begin{table}[t]
\setlength{\tabcolsep}{8pt}
\caption{Properties of different monocular depth estimation datasets. The last column denotes types of hard cases (see Fig.~\ref{fig_abstract_HC}) included in these datasets.
}
\label{tab_method_dataset}
\centering
\scriptsize
{
		\begin{tabular}{|c|c|c|c|c|c|}\hline
			ID & Datasets &Type &Annotation &Images & Hard cases\\ \hline
			1 & NYU& Indoor & Kinect    &108,644  & a,b,c  \\
			2 & ScanNet& Indoor & Kinect &99,571& a,c,f \\
			3 & SUNCG& Indoor &Synthetic    &454,078  & - \\
			4 & CAD& Portrait & Kinect   &145,155& b \\
			5 & URFall & Portrait  & Kinect  &10,764 & b \\ \hline
			6 & HC Depth & Mixed  & Kinect \& RealSense &120,060 & a,b,c,d,e,f \\
			\hline
		\end{tabular}
}
\end{table}

\section{Datasets}
\label{sec:datasets}

\subsection{HC Depth Dataset}

In recent years, several RGBD datasets have been proposed to provide collections of images with associated depth maps.
In Tab.~\ref{tab_method_dataset}, we list five open source RGBD datasets and summarize their properties such as types of content, annotation methods, and number of images.
Among them, NYUDv2~\cite{Silberman:ECCV12}, ScanNet~\cite{dai2017scannet}, CAD~\cite{CAD.org}, and URFall~\cite{kwolek2014human} are captured from real indoor scenes using Microsoft Kinect~\cite{zhang2012microsoft}, while SUNCG~\cite{song2016ssc} is a synthetic dataset collected by rendering manually created 3D virtual scenes.
In addition, CAD and URFall contain videos of humans performing activities in indoor environments.
These datasets offer a large number of annotated depth images and are widely used to train models for an MDE task.
However, each of these RGBD datasets primarily focuses on only one type of scenes and may not cover enough challenging cases.
In Fig.~\ref{fig_abstract_HC}, we identify and summarize six types of typical failure cases for two state-of-the-art MDE methods~\cite{Alhashim2018,fu2018deep} trained on the NYUDv2 dataset.


To complement existing RGBD datasets and provide sufficient coverage on challenging examples for the MDE task, we design and acquire a new dataset, named \emph{HC Depth dataset}, which contains all the six categories of hard cases shown in Fig.~\ref{fig_abstract_HC}.
Specifically, we collect $24660$ images using Microsoft Kinect~\cite{zhang2012microsoft}, which provides dense and accurate depth maps for the corresponding RGB images.
Due to the limited effective distance range of Kinect, these images are mainly about indoor scenes of portraits\nothing{ with slight rotation}.
We also collect $95400$ images of both indoor and outdoor scenes using \nothing{handheld stereoscopic }Intel RealSense~\cite{keselman2017intel}, which is capable of measuring larger depth range in medium precision.
In \textit{sky} cases, we assign a predefined maximum depth value to sky regions based on semantic segmentation~\cite{zhou2018semantic}.
We also perform surface smoothing and completion for all the cases using the toolbox proposed by Silberman \etal~\cite{Silberman:ECCV12}.
We show several typical examples of our HC Depth dataset in the supplementary materials.

\subsection{Incremental Dataset Mixing Strategy}
\label{chap_dataset_combination}
\label{sec:dataset_mixing}

Training on aforementioned datasets together poses a challenge due to the different distributions of depth data in various scenes.
Motivated by curriculum learning~\cite{bengio2009curriculum} in  global optimization of non-convex functions, we propose an incremental \nothing{balanced }dataset mixing strategy to accelerate the convergence of network training and improve the generalization performance of trained models.

Curriculum learning is related to boosting algorithms, in which difficult examples are gradually added during the training process.
In our case of MDE, we divide all the training examples into four main categories based on the content and difficulty, i.e., indoor (I), synthetic (S), portrait (PT), and hard cases (HC).
First, we train our model on datasets with similar distributions (e.g., I + S) until convergence.
Then we add remaining datasets (e.g., PT or HC) one by one and build a new sampler for each batch to ensure a balanced sampling from these imbalanced datasets.
Specifically, we count the number of images $k_i$ contained in each dataset, and $K=\sum_{i}{k_i}$ is the total number of training images.
The probability of sampling an image from the $i$-th dataset is proportional to $K/k_i$ to effectively balance different datasets.
\nothing{
We propose the following multi-stage algorithm to decide sampling probabilities from different datasets:
\begin{description}
\setlength{\itemsep}{0pt}
\setlength{\parskip}{0pt}
\setlength{\parsep}{0pt}
\item[Step 1:] We count the number of images $k_i$ contained in each dataset, and $K=\sum_{i}{k_i}$ is the total number of training images.
\item[Step 2:] The probability of sampling an image from the $i$-th dataset is proportional to $K/k_i$ to effectively balance different datasets.
\item[Step 3:] We multiply the sampling probability by a constant $C_i$ to further control the sampling ratio for better convergence.
\end{description}
}

\section{Experiments}
\label{sec:experiments}

\nothing{
To demonstrate the effectiveness of our method, we conduct a  series of experiments to evaluate different aspects of our approach.
After describe the implementation details, we present our results on NYUDv2 dataset~\cite{Silberman:ECCV12}, zero-shot cross-datasets (TUM)~\cite{sturm2012benchmark}, and our own HC Depth Dataset.
Some ablation studies performed on NYUDv2 are discussed to give more detailed analysis of our method.
}

\subsection{Experimental Setup}
\label{sec:experimental_setup}

To validate each algorithm component, we first train a baseline model $\baselinemodel$ without any module introduced in Sec.~\ref{sec:method}.
We denote the model trained with edge-aware consistency as $\edgeawaremodel$ and the model trained with both edge-aware consistency and our spatial attention blocks as $\sabmodel$.
The model obtained with all the components is denoted as $\fullmodel$, which is essentially $\sabmodel$ trained with the addition of GFRL and $\relativelossmodel$ trained with the addition of SAB.
We also compare with the option to train $\sabmodel$ with an additional relative loss $\relativeloss$ described in Chen \etal~\cite{chen2016single}.
We show additional comparisons with variations of existing attention modules, e.g., Spatial Excitation Block (sSE)~\cite{roy2018concurrent} and Convolutional Block Attention Module (CBAM)~\cite{woo2018cbam}.
Finally, we train the full model on different combinations of datasets to evaluate the effect of adding more training data.

We implement our method using PyTorch\nothing{~\cite{paszke2017automatic}} and train the models on four NVIDIA GPUs with 12GB memory.
We initialize all the ResNet-101 blocks with pretrained ImageNet weights and randomly initialize other layers.
We use RMSProp~\cite{tieleman2012lecture} with a learning rate of $10^{-4}$ for all the layers and reduce the rate by $10\%$ after every $50$ epochs. $\beta_1$, $\beta_2 $ and weight decay are set to $0.9$, $0.99$, and $0.0001$ respectively.
The batch size is set to be $48$.
The weights $\{ \lambda_{i} \}$ of the loss terms described in Sec.~\ref{chap_loss} are set to $1$, $1$, $1$, and $0.5$ respectively.
We pretrain the baseline model $\baselinemodel$ for $40$ epochs on NYUDv2 dataset.
The total number of trainable parameters for our full model is about $167$M.
When training on multiple datasets, we first use our multi-stage algorithm (Sec.~\ref{sec:network_training}) on the first dataset until convergence of the total loss function, and then include more data via our incremental dataset mixing strategy (Sec.~\ref{sec:dataset_mixing}).

\begin{figure}[t]
    \small
    \centering
	\begin{minipage}{.03\linewidth}\centering
		\rotatebox[origin=l]{90}{\makebox{Berhu Loss}}
	\end{minipage}
	\begin{minipage}{.6\linewidth}\centering
		\includegraphics[width=\linewidth]{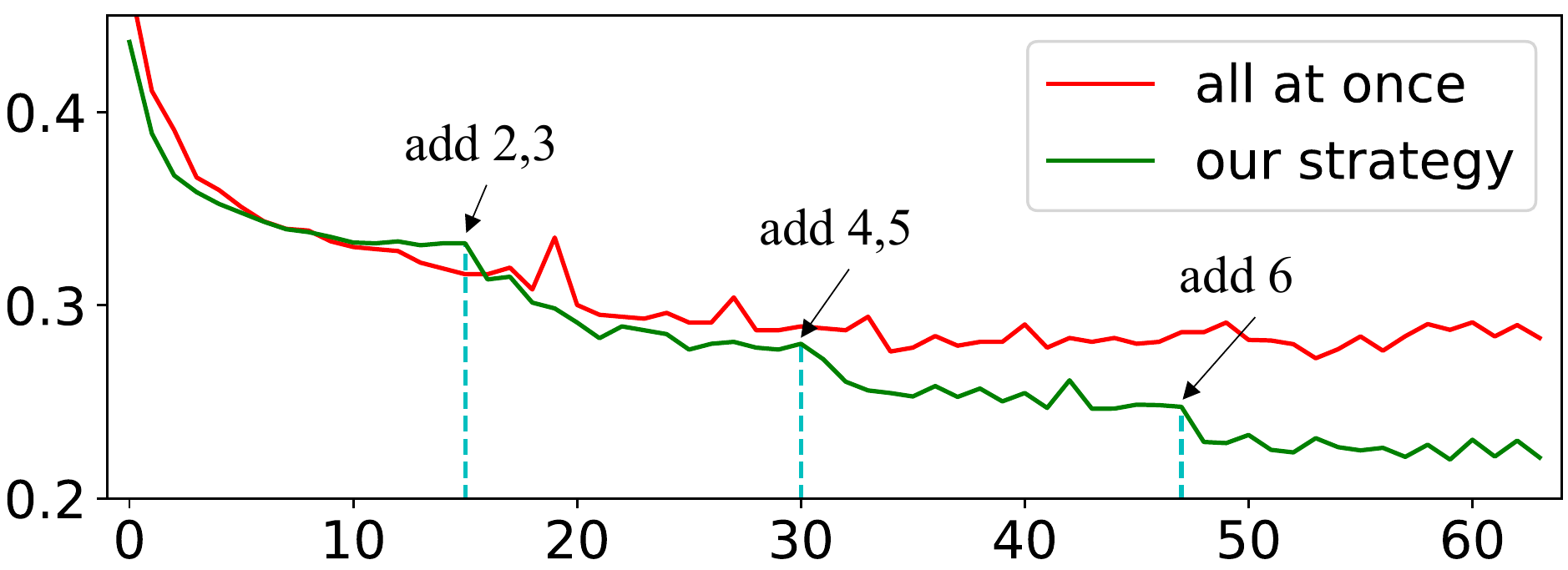}
	\end{minipage}\\
	\begin{minipage}{.06\linewidth}\centering
		\rotatebox[origin=l]{0}{\makebox{Epoch}}
	\end{minipage}
    \caption{BerHu loss curves of different training strategies. The {\color{red} red} curve denotes the loss curve with all datasets added from the beginning, while the {\color{TabGreenColor} green} curve represents the loss curve based on our incremental dataset mixing strategy. The {\color{TabBlueColor} blue} dot lines indicate the epochs when new datasets are included. The dataset IDs are defined in Tab.~\ref{tab_method_dataset}.}
    \label{fig_dataset_convergence}
\end{figure}

\label{chap-metric}
We compare our method with several state-of-the-art MDE algorithms both quantitatively and qualitatively using the following commonly used metrics~\cite{eigen2014depth}:
\begin{itemize}[leftmargin=*]
	\setlength{\parskip}{0pt}
	\setlength{\parsep}{0pt}
	\item Average relative error (REL): $\frac{1}{n} \sum_{p}^{n} \left|d_{p}-\hat{d}_{p}\right|/\hat{d}_{p}$.
	\item Root mean squared error (RMSE):  $\sqrt{\frac{1}{n} \sum_{p}^{n}\left(d_{p}-\hat{d}_{p}\right)^{2}}$.
	\item Average $\text{log}10$ error: $\frac{1}{n} \sum_{p}^{n}\left|\log _{10}\left(d_{p}\right)-\log _{10}\left(\hat{d}_{p}\right)\right|$.
	\item Threshold accuracy: $\{ \accuracy_i \} $ are $\mathrm{ratios \, of \, pixels} \, \{d_{p}\}$ $\mathrm { s.t. }$ $\max \left(\frac{d_{p}}{\hat{d}_{p}}, \frac{\hat{d}_{p}}{d_{p}}\right)<thr_i$ $\mathrm{for}$  $thr_i=1.25,1.25^{2}, \mathrm{and} \, 1.25^{3}$.
\end{itemize}

\subsection{Experimental Results}
\label{sec:experimental_results}
 \begin{figure}[t]
 	\centering
 	\includegraphics[width=0.9\linewidth]{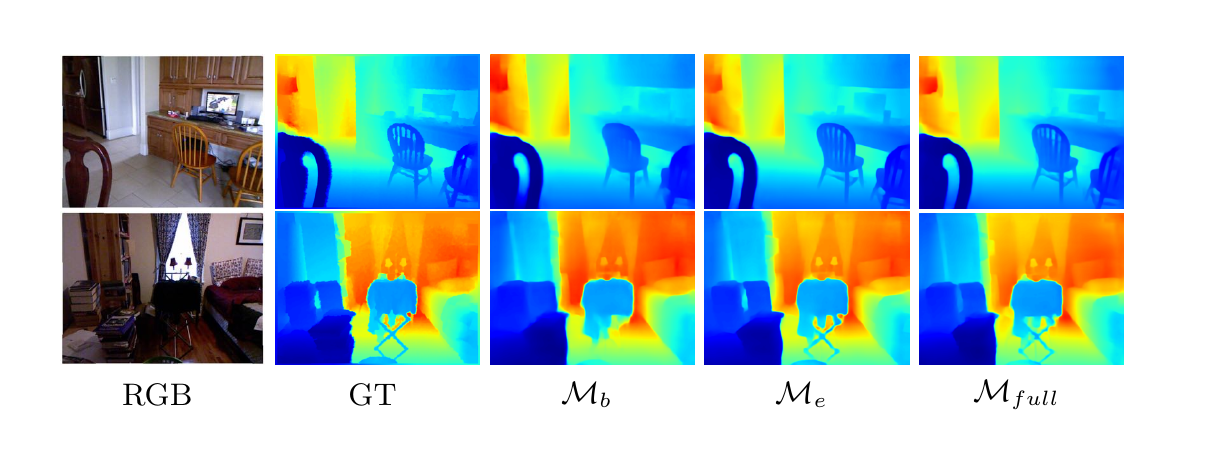}
	\caption{Evaluation results of our edge-aware consistency module. From left to right: input images, ground-truth depth, results of our baseline model $\baselinemodel$, the model $\edgeawaremodel$ with our edge-aware consistency module, and our full model $\fullmodel$.
 	}
 	\label{fig_result_boundary}
 \end{figure}

\begin{table}[t]
\caption{Quantitative results on NYUDv2 dataset using different quantitative metrics.  Higher numbers indicate better accuracy, while lower ones represent better results in terms of predication errors.
The numbers after the right arrow indicate the IDs of datasets (defined in Tab.~\ref{tab_method_dataset}) to train each model.
}
\label{tab_result_nyu}
\scriptsize
{
\setlength{\tabcolsep}{8pt}
	\begin{center}
		\begin{tabular}{|c|ccc|ccc|}\hline
			\multirow{2}{*}{Methods} & \multicolumn{3}{|c|}{ Accuracy $\uparrow$}  & \multicolumn{3}{|c|}{ Error $\downarrow$} \\ \cline{2-7}
			~ &$\accuracy_1$  & $\accuracy_2$  & $\accuracy_3$ & RMSE &REL &$\text{log10}$ \\ \hline
			Eigen \etal~\cite{eigen2015predicting} $\rightarrow$ 1& 0.769 &0.950 &0.988&  0.641 &0.158  &- \\
			Laina \etal~\cite{laina2016deeper} $\rightarrow$ 1  &0.811 & 0.953 & 0.988  & 0.573  &0.127& 0.055 \\
			Qi \etal~\cite{qi2018geonet} $\rightarrow$ 1&0.834 &0.960 &0.990  & 0.569 &0.128   &0.057  \\
			Hao \etal~\cite{hao2018detail} $\rightarrow$ 1 & 0.841 &0.966 &0.991 & 0.555 &0.127 &0.053 \\
			Hu \etal~\cite{hu2019revisiting} $\rightarrow$ 1&0.866  &0.975 &0.993 & 0.530  &0.115	 &0.050 \\
			Fu \etal~\cite{fu2018deep} $\rightarrow$ 1&0.828  & 0.965 &0.992 & 0.509 & 0.115 & 0.051  \\
			Alhashim \etal~\cite{Alhashim2018} $\rightarrow$ 1&0.846  &0.974 &0.994 & 0.465  &0.123 &0.053 \\
			Yin \etal~\cite{yin2019enforcing} $\rightarrow$ 1& 0.875 &	0.976 &	0.994 &	0.416 &	0.108 &	0.048 \\ \hline
			$\baselinemodel$ $\rightarrow$ 1 & 0.856     &0.974  &0.994  & 0.430 &0.120 & 0.051 \\
			$\edgeawaremodel$ $\rightarrow$ 1   & 0.860  &0.974  &0.994  & 0.426 &0.118  &0.050  \\
			$\sabmodel$ $\rightarrow$ 1 &0.864   &0.971  &0.993  & 0.417 &0.113 & 0.049  \\
            $\fullmodel$ $\rightarrow$ 1 & 0.876 &0.979 &0.995 & 0.407 & 0.109 & 0.047  \\ \hline
		    $\sabmodel+\relativeloss$~\cite{chen2016single} $\rightarrow$ 1  & 0.864    &0.971  &0.993  & 0.418 & 0.113 & 0.049  \\
			$\relativelossmodel$+ sSE~\cite{roy2018concurrent} $\rightarrow$ 1 & 0.857 & 0.968 & 0.992 & 0.433 & 0.117 &0.051\\
			$\relativelossmodel$+ CBAM~\cite{woo2018cbam} $\rightarrow$ 1 & 0.860 & 0.968 & 0.992 & 0.432 & 0.117 & 0.050 \\ \hline
			$\baselinemodel$ $\rightarrow$ 1,6 & 0.868 & 0.976 & 0.995 & 0.420 & 0.115 & 0.049 \\
			$\baselinemodel$ $\rightarrow$ 1-6 & 0.874 & 0.978 & 0.995 & 0.414 & 0.111 & 0.048 \\
			$\fullmodel$ $\rightarrow$ 1-3 & 0.888 &0.982 & \textbf{0.996} &0.391 &0.104 &0.044 \\
			$\fullmodel$ $\rightarrow$ 1-5  & 0.888 &0.981 &\textbf{0.996} &0.390 &0.103 &0.044 \\
			$\fullmodel$ $\rightarrow$ 1,6 & 0.885 & 0.979 & 0.994 & 0.401 & 0.104 & 0.045 \\
 			$\fullmodel$ $\rightarrow$ 1-6 & \textbf{0.899} &\textbf{0.983} &\textbf{0.996} & \textbf{0.376} &\textbf{0.098} &\textbf{0.042} \\
			\hline
		\end{tabular}
	\end{center}
}
\end{table}

 \begin{figure}[t]
 	\centering
 	\includegraphics[width=.9\linewidth]{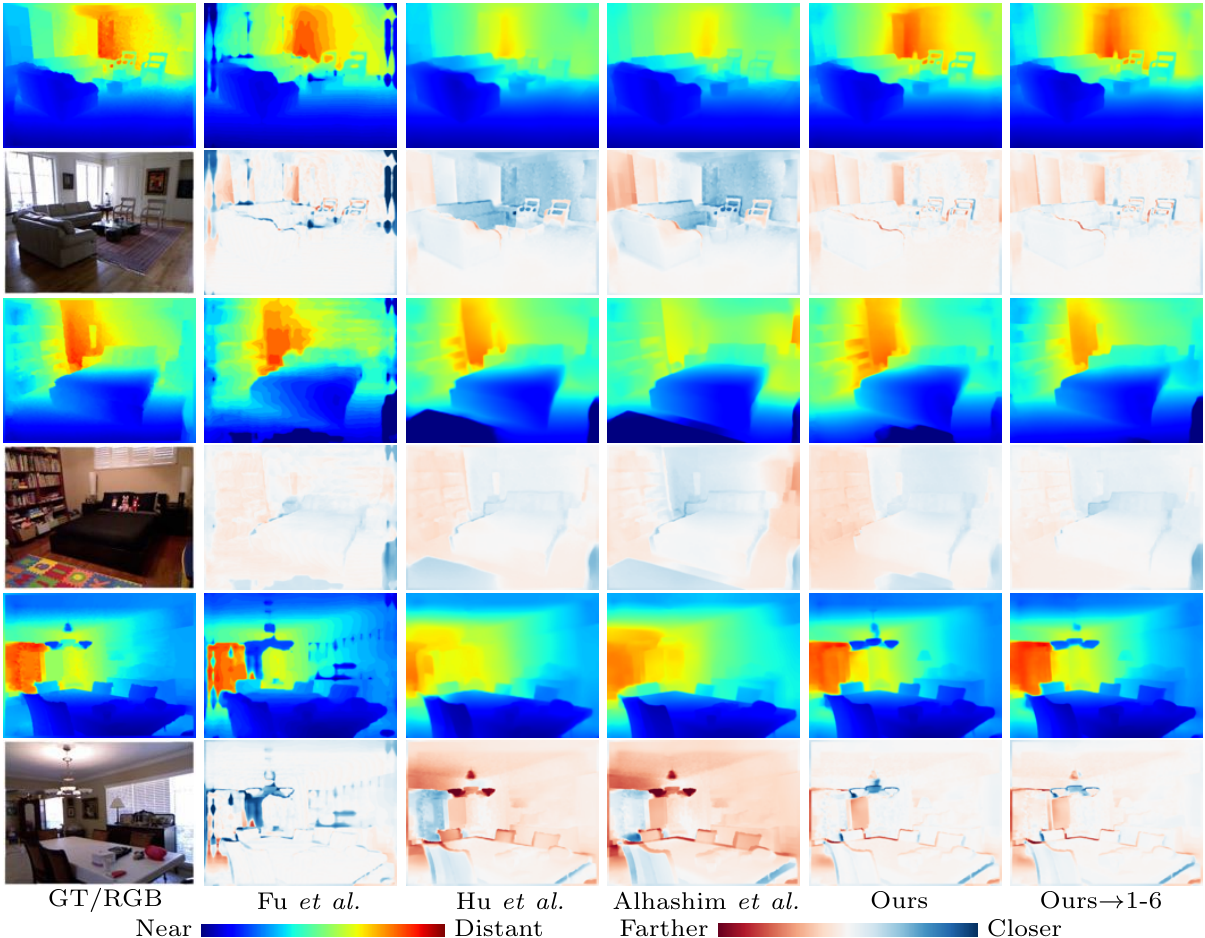}
 	\caption{Qualitative results on four test cases from NYUDv2 Dataset.
 		In each example, the first row shows the ground-truth depth map, results of three prior methods~\cite{fu2018deep,hu2019revisiting,Alhashim2018}, and the results of our full model trained solely on NYUDv2 and on the combination of all the six datasets listed in Tab.~\ref{tab_method_dataset}, respectively.
 		The second row of each example shows the input image and the error maps of the corresponding results in the first row.}
 	\label{fig_result_nyu}
 \end{figure}

\subsubsection{Results on NYUDv2 Dataset.}
The NYUDv2 dataset contains $464$ indoor scenes.
We follow the same train/test split as previous work, i.e., to use about 50K images from $249$ scenes for training and $694$ images from 215 scenes for testing.
In Fig.~\ref{fig_dataset_convergence}, we compare the BerHu loss curves on the test set when training on multiple datasets \emph{without} and \emph{with} our incremental dataset mixing strategy (Sec.~\ref{sec:dataset_mixing}), which illustrates that our strategy considerably improves the convergence of network training.

Tab.~\ref{tab_result_nyu} summarizes the quantitative results of our ablation study on NYUDv2 dataset together with the numbers reported in previous work.
Our baseline model $\baselinemodel$ uses ResNet101 as the backbone and couples with GCB modules, combining several widely used loss terms (Berhu loss, scale-invariant gradient loss, and normal loss) of previous work.
As can be seen from the table, 
all of our algorithm components can improve depth estimation results noticeably, including edge-aware consistency, SAB, and GFRL.
Furthermore, adding our HC Depth dataset with our dataset mixing strategy (Sec.~\ref{sec:dataset_mixing}) significantly improves the model performance (see $\baselinemodel$ $\rightarrow$ 1,6 and $\fullmodel$ $\rightarrow$ 1,6).
By comparing $\fullmodel$ with  $\sabmodel+\relativeloss$~\cite{chen2016single}, we can conclude that GFRL leads to better results than the alternative relative loss~\cite{chen2016single}.
We also show that our SAB ($\fullmodel$) brings notable improvement over Spatial Excitation Block ($\relativelossmodel$+ sSE~\cite{roy2018concurrent}) and Convolutional Block Attention Module ($\relativelossmodel$+ CBAM~\cite{woo2018cbam}).
Finally, training our full model $\fullmodel$ on multiple datasets can achieve the best results and outperforms training solely on NYUDv2 dataset by a large margin.

Fig.~\ref{fig_result_nyu} presents qualitative results on three test cases to compare our full model with three prior methods~\cite{fu2018deep,hu2019revisiting,Alhashim2018}.
The first example shows that our results have more reasonable spatial layout.
In the second example, our model provides more accurate estimation than other methods in the regions of color boundaries.
In the third example, our results preserve details in the region around the chandelier.
Although Fu \etal~\cite{fu2018deep} achieves structural patterns similar to our method, their results contain disordered patterns and thus have much larger errors.
Fig.~\ref{fig_result_boundary} shows evaluation results on two more test examples from NYUDv2 to compare our baseline model $\baselinemodel$ with the model $\edgeawaremodel$ trained with edge-aware consistency and our full model $\fullmodel$.
Our edge-aware consistency module leads to much more accurate boundaries in the depth estimation results, such as the back of the chair and legs of the desk.
 \begin{figure}
 	\centering
 	\includegraphics[width=.9\linewidth]{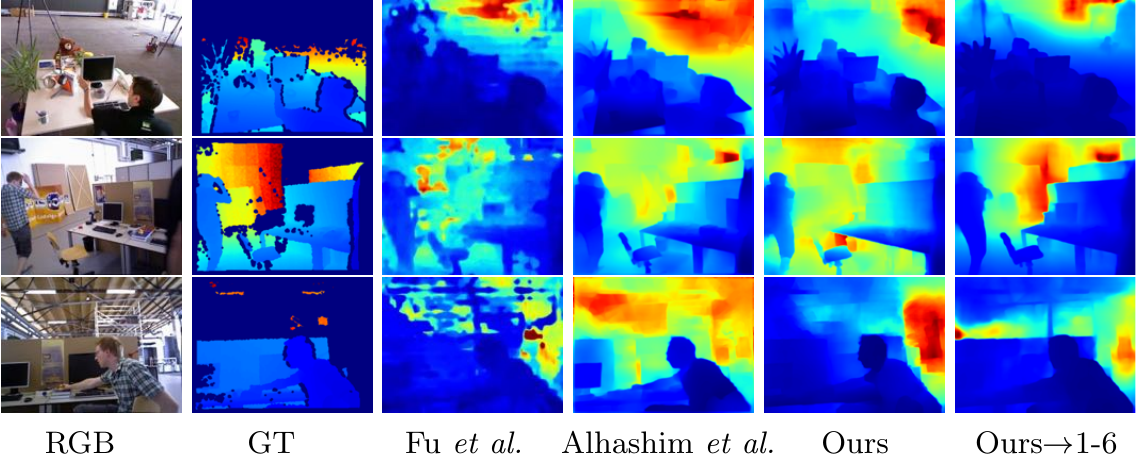}
     \caption{Qualitative results on TUM dataset.}
 	\label{fig_result_tum}
 \end{figure}

\begin{table}
\caption{Quantitative results of generalization test on TUM dataset. The experimental configuration is the same as Tab.~\ref{tab_result_nyu}.}
\label{tab_result_tum}
\scriptsize
{
\setlength{\tabcolsep}{8pt}
	\begin{center}
		\begin{tabular}{|c|ccc|ccc|}\hline
			\multirow{2}{*}{Methods} & \multicolumn{3}{|c|}{ Accuracy $\uparrow$} & \multicolumn{3}{|c|}{ Error $\downarrow$} \\ \cline{2-7}
			~ &$\accuracy_1$ & $\accuracy_2$  & $\accuracy_3$ & RMSE &REL&$\text{log10}$ \\ \hline
			Hu \textit{et al.}~\cite{hu2019revisiting} $\rightarrow$ 1 &0.577  &0.842 &0.932 & 1.154  &0.216 &0.111 \\
			Fu \textit{et al.}~\cite{fu2018deep} $\rightarrow$ 1&0.598  & 0.855 &0.934 & 1.145 & 0.209 & 0.110  \\
			Alhashim \textit{et al.}~\cite{Alhashim2018} $\rightarrow$ 1&0.567  &0.847 &0.920 & 1.250  &0.224 &0.115 \\
			\hline
			$\fullmodel \rightarrow$1 & 0.606      & 0.888  & 0.947  &1.109   &0.212  &0.100 \\
			$\fullmodel \rightarrow$ 1-3         &   0.665	    &0.903  &0.955 &1.031 &0.194 &0.091  \\
			$\fullmodel \rightarrow$ 1-5        & 0.710       &0.913   & 0.952  &1.013    &  \textbf{0.175}   & 0.084  \\
		    $\fullmodel \rightarrow$1,6 & 0.689 & 0.906 & 0.950 & 1.029 & 0.189 & 0.086 \\
 			$\fullmodel \rightarrow$ 1-6 & \textbf{0.735}	    &\textbf{0.927}   & \textbf{0.959}   & \textbf{0.912}  &  0.177  & \textbf{0.081}   \\ \hline
		\end{tabular}
	\end{center}
}
\end{table}

 \begin{figure}
 	\centering
 	\includegraphics[width=.9\linewidth]{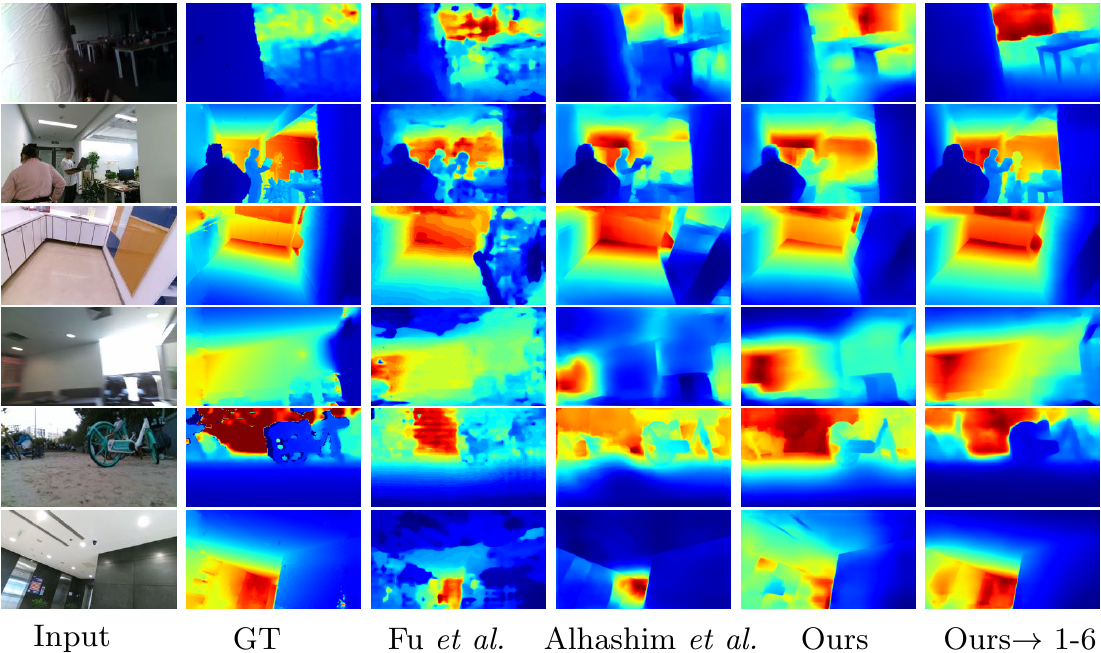}
 	\caption{Qualitative results on our HC Depth dataset.
 	From top to bottom, the examples correspond to the six types of hard cases showed in Fig.~\ref{fig_abstract_HC}.
    }
 	\label{fig_result_HC}
 \end{figure}

\begin{table}[t]
\caption{Quantitative results on the HC Depth dataset. The experimental configuration is the same as Tab.~\ref{tab_result_nyu}.}
\label{tab_result_HC}
\scriptsize
{
\setlength{\tabcolsep}{8pt}
	\begin{center}
		\begin{tabular}{|c|ccc|ccc|}\hline
			\multirow{2}{*}{Methods} & \multicolumn{3}{|c|}{ Accuracy $\uparrow$}  & \multicolumn{3}{|c|}{ Error $\downarrow$} \\ \cline{2-7}
			~ &$\accuracy_1$ & $\accuracy_2$  & $\accuracy_3$ & RMSE &REL&$\text{log10}$ \\ \hline
			Hu \textit{et al.}~\cite{hu2019revisiting} $\rightarrow$ 1 &0.531  &0.783 &0.898 & 1.276  &0.285 &0.128 \\
			Fu \textit{et al.}~\cite{fu2018deep} $\rightarrow$ 1&0.477  & 0.755 &0.866 & 1.356 & 0.2962& 0.145  \\
			Alhashim \textit{et al.}~\cite{Alhashim2018} $\rightarrow$ 1&0.551  &0.819 &0.918 & 1.137  &0.257 &0.118 \\
			\hline
			$\fullmodel \rightarrow$1& 0.600  &0.843  &0.930 & 1.070 &0.249 &0.109 \\
			$\fullmodel \rightarrow$ 1-3 & 0.610 & 0.837  & 0.921  &1.072   &0.244  &0.108  \\
			$\fullmodel \rightarrow$ 1-5 & 0.633   &0.845   & 0.924  &1.065    &  0.237   & 0.107  \\
			$\fullmodel \rightarrow$1,6 & 0.825 & 0.895 & 0.961 & 0.715 & 0.190 & 0.087 \\
			$\fullmodel \rightarrow$ 1-6 &\textbf{0.879}   &\textbf{0.965}  & \textbf{0.988}   & \textbf{0.566}  &  \textbf{0.113}  & \textbf{0.048}   \\ \hline
		\end{tabular}
	\end{center}
}
\end{table}

\subsubsection{Results on TUM Dataset.}
We use the open source benchmark TUM~\cite{sturm2012benchmark} to evaluate the generalization performance of different methods\nothing{ and models} in the setting of zero-shot cross-dataset inference.
The test set of TUM consists of $1815$ high quality images taken in a factory including pipes, computer desks and people, which are never seen when we train our models.
Tab.~\ref{tab_result_tum} demonstrates that our full model $\fullmodel$ trained solely on NYUDv2 dataset outperforms previous methods.
Furthermore, training the full model by adding HC Depth dataset or more datasets using our dataset mixing strategy can significantly improve the generalization performance.
As shown in the qualitative results in Fig.~\ref{fig_result_tum}, our method retains sharp edges of plants in the first example and leads to more accurate spatial layout in the second and third examples.

\subsubsection{Results on HC Depth Dataset.}
To test the performance of different models on hard cases, we use the test split of our HC Depth dataset which contains $328$ examples.
Tab.~\ref{tab_result_HC} summarizes the corresponding quantitative results, from which we can obtain consistent findings with Tab.~\ref{tab_result_tum}.
Fig.~\ref{fig_result_HC} illustrates qualitative results of six examples from the test set, which correspond to the six types of hard cases in Fig.~\ref{fig_abstract_HC}.
As shown in Fig.~\ref{fig_result_HC} and Tab.~\ref{tab_result_HC}, our method based on NYUDv2 dataset already provides more faithful prediction compared to Alhashim \etal~\cite{Alhashim2018} and the predicted depth can be further improved considerably by adding our HC Depth dataset or using all the datasets through our dataset mixing strategy.

\section{Conclusions}
\label{sec:conclusions}
In this paper we put together a series of coherent efforts to improve the structural awareness in monocular depth estimation, with the effectiveness and necessity of each component thoroughly verified.
We introduce a novel encoder-decoder architecture using the spatial attention mechanism, and boost the network performance by proposing a global focal relative loss and an edge-aware consistency module\nothing{ to train the model}.
We further collect a dataset of hard cases for the task of depth estimation and leverage a data mixing strategy based on curriculum learning for effective network training.
We validate each component of our method via comprehensive ablation studies and demonstrate substantial advances over state-of-the-art approaches on benchmark datasets.
Our experimental results show that truly generic models for monocular depth estimation require not only innovations in network architecture and training algorithm, but also sufficient data for various scenarios.

Our source code, pretrained models, and the HC Depth dataset will be released to encourage follow-up research.
In the future, we plan to capture more diverse scenes and further expand our HC Depth dataset.
We would also like to 
deploy our models on mobile devices for several applications such as augmented reality.

\subsubsection{Acknowledgements.}
We would like to thank the anonymous reviewers for their valuable comments, Jiwen Liu for help on preparing our dataset, and Miao Xuan for help on paper proofreading.


\bibliographystyle{splncs04}
\bibliography{refsm}

\end{document}